\documentclass[12pt, a4paper]{elsarticle}

\usepackage{amssymb}
\usepackage{amsmath}
\usepackage[ruled,vlined]{algorithm2e}
\usepackage{graphicx}
\usepackage{subcaption} % or \usepackage{subfigure}
\usepackage[export]{adjustbox}
\usepackage{float}
\usepackage{hyphenat}

\restylefloat{table}

\begin{document}

\begin{frontmatter}



\title{Latent Space Perspicacity and Interpretation Enhancement (LS-PIE) Framework}


\author{Jesse Stevens*, Daniel N. Wilke, Itumeleng Setshedi}
\address{Department of Mechanical and Aeronautical Engineering, University of Pretoria, Lynnwood Rd, Hatfield, Pretoria, 0086}

\begin{abstract}
Linear latent variable models such as principal component analysis (PCA), independent component analysis (ICA), 
canonical correlation analysis (CCA), and factor analysis (FA) identify latent directions (or loadings) either ordered or unordered. The data is then projected onto the latent directions to obtain their projected representations (or scores). For example, PCA solvers usually rank the principal directions by explaining the most to least variance, while
ICA solvers usually return independent directions unordered and often with single sources spread across multiple directions as multiple sub-sources, which is of severe detriment to their usability and 
interpretability. 

This paper proposes a general framework to enhance latent space representations for improving the interpretability of 
linear latent spaces. Although the concepts in this paper are language agnostic, the framework is written in Python.
This framework automates the clustering and ranking of latent vectors to enhance 
the latent information per latent vector, as well as, the interpretation of latent vectors.
Several innovative enhancements are incorporated including latent ranking (LR), latent scaling (LS), 
latent clustering (LC), and latent condensing (LCON).

For a specified linear latent variable model, LR  ranks latent directions according to a specified metric, 
LS scales latent directions according to a specified metric, LC automatically clusters latent
directions into a specified number of clusters, while, LCON automatically determines  
an appropriate number of clusters into which to condense the latent directions for a given metric. Additional functionality of the framework includes single-channel 
and multi-channel data sources, data preprocessing strategies such as Hankelisation to seamlessly expand the applicability of linear latent variable models (LLVMs) 
to a wider variety of data.

The effectiveness of LR, LS, and LCON are showcased on two crafted foundational problems with two applied latent variable models, 
namely, PCA and ICA.

\end{abstract}


\end{frontmatter}

\section{Introduction}

Latent variable models are statistical models that aim to describe the relationships between observed variables and unobserved, or latent, variables. 
These models assume that the observed variables are generated by the underlying latent variables, which are not directly measured or observed but
are inferred from available data \cite{booyse, wilke2022a}. Practically, latent variable models (LVMs) can be classified into reconstruction- and interpretation-centered models
\cite{wilke2022}.

Reconstruction-centered LVMs identify compressed latent representations that are efficient in reconstructing the variance
in the data, often optimal for the given model flexibility. In turn, interpretation-centered LVMs attempt to identify latent presentations 
that are interpretable, e.g. independent variance contributing sources, when explaining the variance in the data. The latter often results in lesser 
compressed latent representations. 

The tasks of these two approaches are distinct as reconstruction-centered models are efficient at compressing data into lower dimensional latent spaces for efficient reconstruction,
while interpretation-centered approaches aim to identify lower-dimensional latent spaces that are interpretable, where contributing factors or sources of variance in 
the data are independent and untangled \cite{wilke2022}.

Ironically, reconstruction-centered LVMs usually present their latent directions
ordered from explaining the most to least variance, or vice versa. These include singular value decomposition (SVD)  \citep{Lever2017}\cite{ZakariaJaadi2022}, principal component analysis (PCA), and, conventional singular spectrum analysis (SSA), 
giving clear discernability to reconstruction-focussed latent representations, while interpretation-centered LVMs usually return the latent directions unordered, with 
single sources spread across multiple directions, making these latent representations less informative and more difficult to discern, interpret and manage. 
These include independent component analysis (ICA) \cite{Tharwat2018}\cite{DeLathauwer2000}\cite{Hyvarinen}, with a variety of underlying objective functions to be maximised such as non-Guassianity measures or proxies such as negentropy, skewness, or kurtosis, or minimisation of mutual information between latent variables.

Our proposed framework addresses this discrepancy by enhancing latent space representations to improve the interpretability of 
linear or locally linearised latent spaces. This framework automates the clustering and ranking of latent vectors according to user-specified metrics 
to improve interpretability and enhance the latent information. 

\begin{figure}[ht!]
\centering
\includegraphics[scale=0.65]{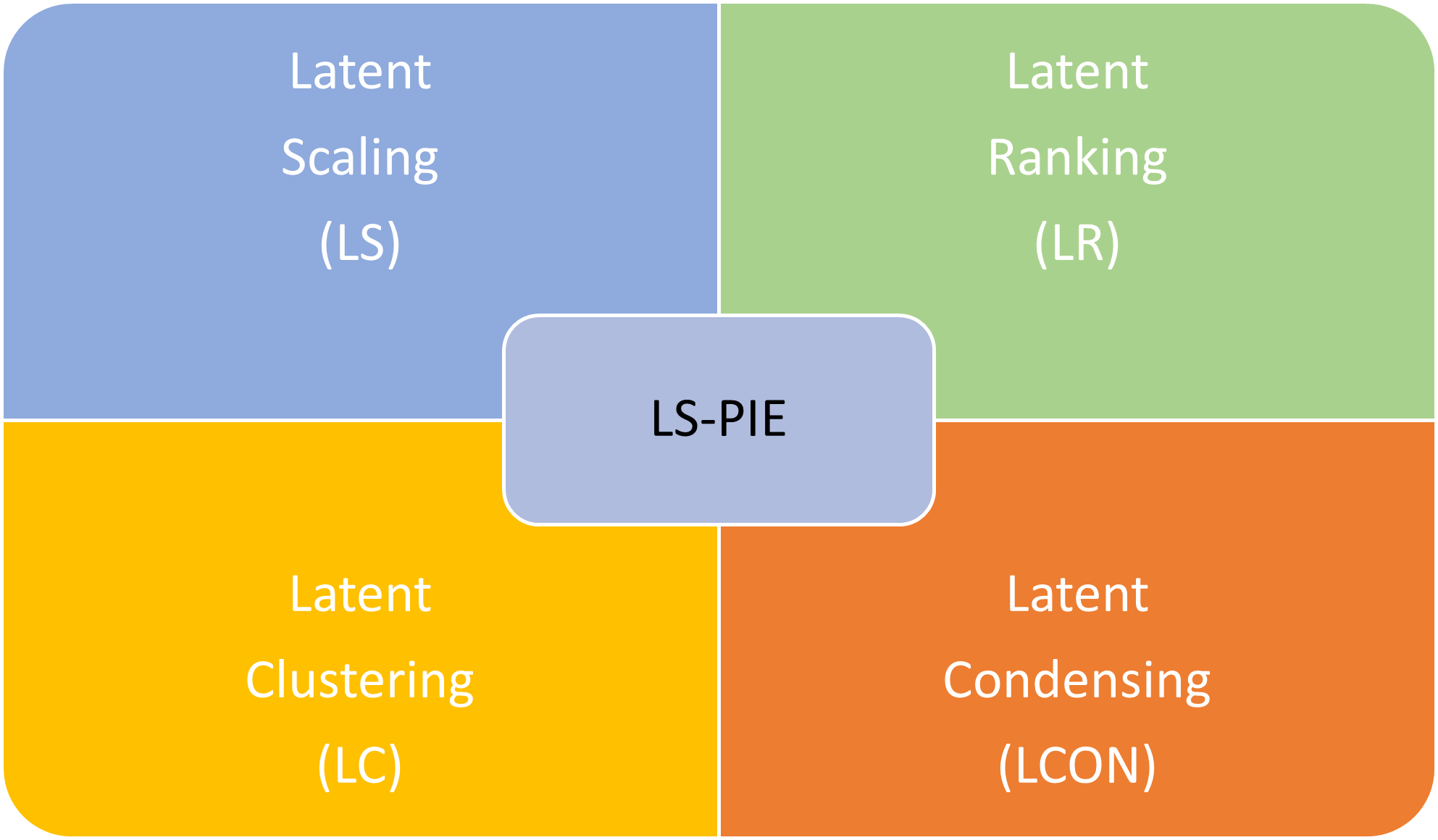}
\linebreak
\small
\caption{LS-PIE Framework highlighting four foundational technologies to enhance latent representations.}\label{fig:lspie4}
\end{figure}

Several innovative enhancements are incorporated including latent clustering (LC), 
latent ranking (LR), and latent condensing (LCON) shown in Figure~\ref{fig:lspie4}. Enhancements can be applied to latent spaces resulting from reconstruction-centered and interpretation-centered LVMs,
to re-rank already ordered latent variables according to an alternative metric, to order unordered latent variables, or to interrogate the influence of pre-processing 
or filtering of data on latent interpretability 
\cite{Westad2005}, to mention some use cases. All of these cases have significant practical and research implications.

The effectiveness of LR, LS, LC, and LCON are show-cased on two crafted foundational problems for two applied latent variable models, 
namely, PCA and ICA, respectively representative of reconstruction-centered and interpretation-centered LVMs.

\section{Required Background}

\subsection{Latent Variable Models}

\begin{figure}[ht!]
\centering
\includegraphics[scale=0.45]{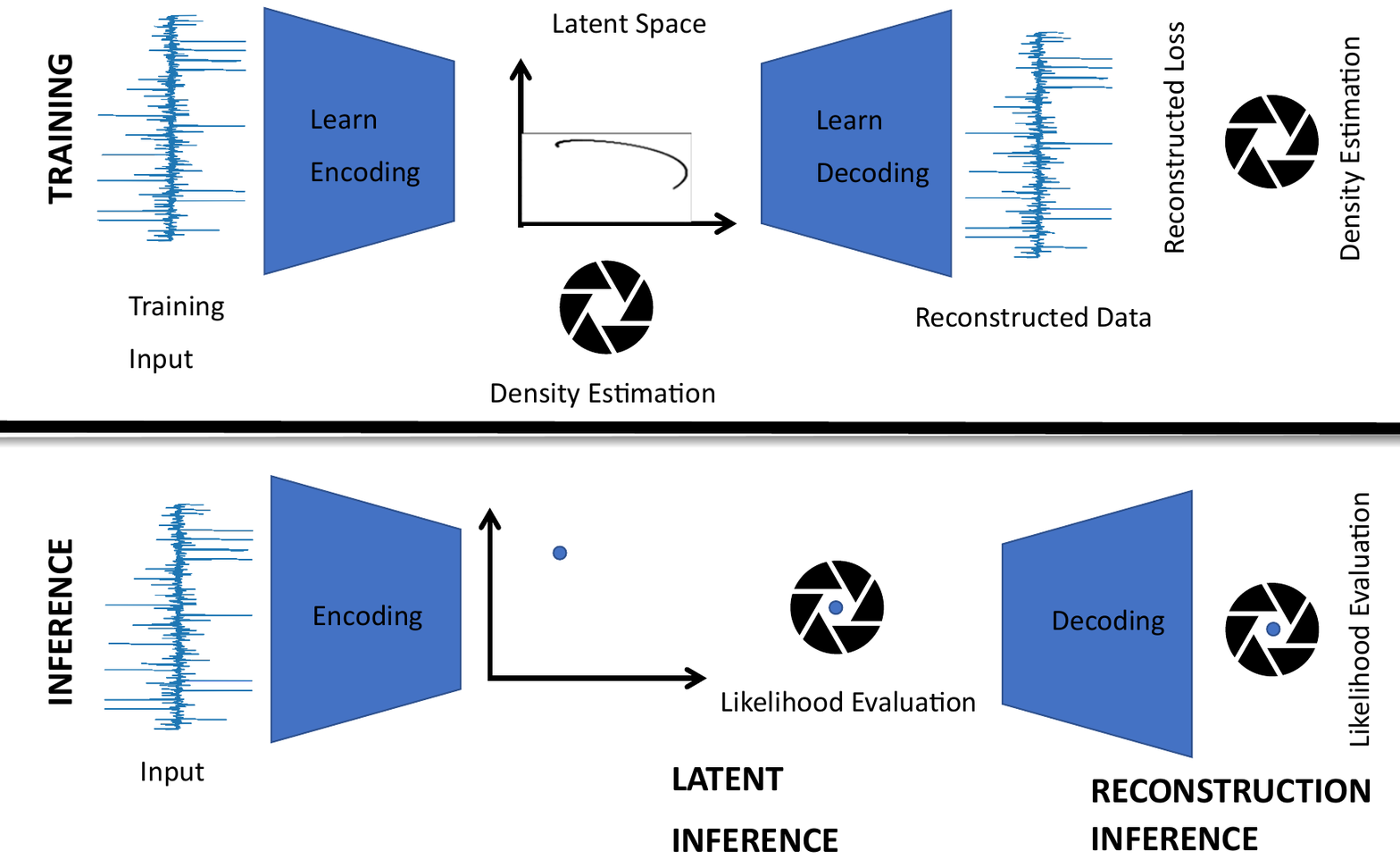}
\linebreak
\small
\caption{The auto-encoding structure is often used in latent variable modeling (LVM). The training stage fits the model on the provided data by learning the weights for encoding and decoding. A higher-dimensional data sample is encoded 
to a lower-dimensional latent space by encoding. The latent representation can then be decoded to obtain a higher-dimensional reconstruction of the sample from its latent representation through decoding. These models are commonly used in many fields of data science for the purpose of data compression and reconstruction, while latent inferencing is gaining traction recently \cite{booyse}.}\label{fig:lvm}
\end{figure}

Auto-associative \cite{kohonen1974}, or auto-encoding \cite{rumelhart1986} is a fundamental concept in LVMs to make the unsupervised learning problem of finding an appropriate latent representation tractable. A conceptual outline of auto-encoding is shown in Figure~\ref{fig:lvm}, depicting encoding and decoding. Encoding transforms higher-dimensional input data into a lower-dimensional latent representation
while decoding transforms the latent representations of higher-dimensional data back to their higher-dimensional representations. Variance-driven LVMs compress input data into compact latent representations, 
while source-driven latent representations aim to identify informative latent representations that indicate sources contributing to the variance in the data.

Inferencing on latent representations or reconstructed representations enables latent and reconstruction inferencing, respectively. 
The aim of this framework is to enhance the latent representations of LVMs for improved and enhanced latent inferencing.

Given a set of time series data represented by an $m \times n$ matrix $\boldsymbol{\bar{X}}$, where $m$ is the number of observations and $n$ is the number of variables or the discrete times at which data is recorded. For time series data the former relates to the number of samples, while
the latter is the time length of each sample. Latent variable models typically proceed as follows:
\begin{enumerate}
\item Data standardisation through mean centering or whitening of the data, $\boldsymbol{X}$.

\item Compute the $n \times n$ covariance matrix $\boldsymbol{C}$ of the standardized time series data $\boldsymbol{X}$.

\item Find latent directions for the data $\boldsymbol{C}$ by maximising or minimising an objective function plus regularisation terms subject to equality\footnote{Equality constraints between latent directions are often enforced 
e.g. orthogonality of the latent directions or some transformed representation of the latent directions. This can always be solved by direct optimisation but solving the first-order necessary optimality condition, or a matrix decomposition may in some cases be computationally more efficient \cite{snyman2018practical}.} and inequality constraints \cite{snyman2018practical}. PCA diagonalises the covariance matrix by finding its eigenvectors and eigenvalues, where the eigenvectors represent the principal components, and the eigenvalues represent the amount of variance explained by each principal component.

\item Select $k$ latent directions from a maximum of $rank(\boldsymbol{C})$. Eigenvalues and their associated eigenvectors are automatically sorted in descending order, from which $k$ eigenvectors associated with the largest eigenvalues are usually selected. By choosing eigenvectors corresponding to the largest eigenvalues the LVM prioritizes reconstruction as they capture the most significant variation in the time series data.

\item The latent representation for a sample is obtained by projecting the standardised time series data $\boldsymbol{C}$ onto $k$ selected latent directions (a.k.a. loadings) to obtain a $k$-dimensional latent representation of the sample often referred to as a $k$-dimensional score.

\item Reconstructing the sample from the latent representation, merely requires the summation of each component of the $k$-dimensional score multiplied by their respective latent direction.
\end{enumerate}

In this paper, we consider sklearn's PCA as a representative reconstruction-centered LVM, and independent component analysis (ICA) in the form FastICA, as an interpretation-centered LVM.
 
\subsection{Data sources and channels}

In data science, there are various sources of data that data analysts and scientists process to gain insights and make informed decisions.

Time series data refers to a sequence of observations or measurements taken at specific time intervals. Time series data is often collected 
from sensors installed in various devices or environments. This can include temperature readings, air quality measurements, pressure recordings, 
vibration data, and more. Industries such as manufacturing, energy, and environmental monitoring heavily rely on sensor-generated time series data.
The Internet of Things (IoT) has introduced a wide range of devices that are now equipped with sensors and connected to the internet. These devices generate time series data that can be 
used for applications like smart homes, smart cities, and industrial monitoring. Medical devices, wearable devices, and health monitoring systems sense heart rates, blood pressure, glucose levels, 
sleep patterns, and other physiological measurements, which aid healthcare analysis, disease detection, and personalised medicine.

Datasets can be single or multi-channel sensor measurements of single or multiple observations. These multi-channel or multiple observations of data can be homogenous or heterogeneous. 
A single observation of a single channel time series data $\mathbf{x} \in R^{m+n-2}$ can be transformed to enable LVMs to operate on the data. These include
Hankelisation \cite{BROOMHEAD1986}
\[
\boldsymbol{H} =
\begin{bmatrix}
x_0 & x_1 & x_2 & \cdots & x_{n-1} \\
x_1 & x_2 & x_3 & \cdots & x_n \\
x_2 & x_3 & x_4 & \cdots & x_{n+1} \\
\vdots & \vdots & \vdots & \ddots & \vdots \\
x_{m-1} & x_m & x_{m+1} & \cdots & x_{m+n-2}, \\
\end{bmatrix}
\]
while multi-observation or multi-channel sources can be considered in isolation or also transformed to enhance latent inferencing.

\section{Related Work}

Clustering methods have been proposed to improve the efficacy of ICA \citep{Bach2003} by using Tree-dependant Component Analysis (TCA). The TCA combines graphical models and Gaussian stationary contrast function to derive richer dependency classes. ICA and clustering have mainly focused on the use of ICA in pattern recognition and image classification analysis, while Expectations Maximisation (EM), K-Means, and fuzzy C-Means have shown satisfactory results when applied to imaging \citep{Widom1966,Yao2012,Zhao2009}. In contrast, our proposed LS-PIE framework introduces a generic framework for the enhancement of LVMs through
latent ranking (LR), latent scaling (LS), latent clustering (LC), and latent condensing (LCON).

\section{Software description}
\label{}
LS-PIE makes latent ranking (LR), latent scaling (LS), latent clustering (LC), and latent condensing (LCON) accessible for reconstruction-centered or interpretation-centered LVMs.
An outline of LR, LS, LC, and LCON is given to conceptualising the approaches followed in each.

\subsection{Latent ranking (LR)}

Latent ranking (LR) allow the user to specify a metric and then rank the latent variables according to the selected metric. Although several metrics are
readily available, the framework also allows for a metric specified as a user-defined Python function. LR algorithm is outlined in Algorithm~\ref{alg:lr}.
\begin{algorithm}[htb!]
\caption{Latent Ranking (LR)}
\KwData{Latent components $L_i,\;1,\dots,M$ and user-selected or user-specified ranking metric $R$ with associated ranking operator $\mathcal{R}$}
\KwResult{Scaled latent components $\tilde{L}_i,\;1,\dots,M$.}
\Begin{
\For{each $\tilde{L}_i,\;1,\dots,M$}
{
    $$\mathcal{L}_i,\;1,\dots,M  = \mathcal{R}(\tilde{L}_i,\;1,\dots,M),$$
    where $R(\mathcal{L}_i+1) > R(\mathcal{L}_i),\;i=1,\dots,M-1$.
}
}
\label{alg:lr}
\end{algorithm}

Latent ranking allows for the exploration
latent variables that have already been identified by optimising some regularised objective function subject to constraints. Also, this enables unordered latent variables to be ordered or 
ordered latent variables to be ordered according to some other metric that enhances the interpretation of the current latent variables or explores some of their underlying characteristics.

\subsection{Latent scaling (LS)}

Latent scaling (LS) allows the user to specify a metric by which to scale the length of the latent vectors, e.g.\ the percentage variance explained. This enhances the 
visual interpretation of the latent vectors when plotted to critically interrogate them. The framework supports a number of metrics but also allows for metrics expressed as user-defined Python functions. 
The LS algorithm is outlined in Algorithm~\ref{alg:ls}.

\begin{algorithm}[htb!]
\caption{Latent Scaling (LS)}
\KwData{Latent components $L_i,\;1,\dots,M$ and user-selected or user-specified scaling metric $s_i$ and associated scaling operator $S_i$}
\KwResult{Scaled latent components $\tilde{L}_i,\;1,\dots,M$.}
\Begin{
\For{each $L_i$}
{
    $$\tilde{L}_i  = S_i(L_i)$$
}
}
\label{alg:ls}
\end{algorithm}

A practical example is to generate a scaling score, \[ s_j = \frac{\theta_j}{ \sum_{i = 1}^M \theta_i}, \] for each $j^\textrm{th}$ latent direction using some metric $\theta$, with
associated scaling operator \[\tilde{L}_j = S_j(L_j) = L_j/s_j.\]
This allows for the scaling of each latent direction based on the scaling score. Scaling scores include variance explained and kurtosis. This essentially allows us to highlight latent directions that a prominent in
reconstruction, while hiding latent directions that do not contribute significantly to reconstruction.

\subsection{Latent clustering (LC)}

We also need to counter an additional problem common in some LVMs. Unlike PCA, where as the number of components increases
the existing components remain unchanged and new components explain less and less of the variance, some LVMs split the same information over
more and more components as the number of latent directions increases, e.g. ICA. Latent clustering combines
similar latent directions according to a user defined metric into a single latent direction through clustering.

For linear models, the maximum number of latent dimensions is dictated by the rank of the data matrix $\boldsymbol{X}$ after standardisation.
Latent clustering (LC) enables the user to specify the number of latent clusters to be identified from the specified number of latent variables, i.e. LC clusters latent directions into a pre-selected number of clusters. 
LC can be performed by selecting available similarities or 
dissimilarity metrics, or as a user-specified Python function, as well as the clustering approach with BIRCH being the default for LC.

\begin{algorithm}[htb!]
\caption{Latent Clustering (LC)}
\KwData{Matrix $\boldsymbol{X}\in \mathcal{R}^{m\times n}$, with $m<n$}
\KwResult{User-specified $K\in\mathcal{N}$ latent components clusters $\bar{L_k},\;1,\dots,K$ from  $M\in\mathcal{N}$ latent components $L_i,\;1,\dots,M$, with $K<M$.}
\Begin{
LVM decomposition of $m\times n$ matrix $\boldsymbol{X}$ into $M$ latent directions $L_i,\;1,\dots,M$. \\
Find $K < M$ clusters $C_k,\;k=1,\dots,K$ using appropriate clustering algorithm, e.g. BIRCH\\
\For{each $C_k$}
{
    $$\bar{L}_k  = \sum_{ j \in C_k} L_j.$$
}
\If{FILTER is TRUE}{
Apply a user-specified filter to $\bar{L}_k$, with the default being a lowpass Butterworth filter.
}
}
\label{alg:lc}
\end{algorithm}

\subsection{Latent CONdensing (LCON)}

Latent condensing extends LC by automatically finding the optimal number of latent dimensions into which to express the latent directions using an 
appropriate clustering algorithm with DBSCAN being the default for LCON.

\begin{algorithm}[tbh!]
\caption{Latent CONdensing (LCON)}
\KwData{Matrix $\boldsymbol{X}\in \mathcal{R}^{m\times n}$, with $m<n$}
\KwResult{Automically identified $K\in\mathcal{N}$ latent components clusters $\bar{L_k},\;1,\dots,K$ from  $M\in\mathcal{N}$ latent components $L_i,\;1,\dots,M$, with $K \leq M$.}
\Begin{
LVM decomposition of $m\times n$ matrix $\boldsymbol{X}$ into $M$ latent directions $L_i,\;1,\dots,M$. \\
Automatically identify $K < M$ clusters, $C_k,\;k=1,\dots,K,$ using appropriate clustering algorithm, e.g. DBSCAN\\
\For{each $C_k$}
{
    $$\bar{L}_k  = \sum_{ j \in C_k} L_j.$$
}
\If{FILTER is TRUE}{
Apply a user-specified filter to $\bar{L}_k$, with the default being a lowpass Butterworth filter.
}

}
\label{alg:lcon}
\end{algorithm}

\section{Numerical investigation}

The effectiveness of LR, LS, and LCON are showcased on two crafted foundational problems using single channel data. In both cases Hankelisation is employed before applying two latent variable models, 
namely, PCA and ICA.

\subsection{Single Channel - Latent ranking (LR), latent scaling (LS), and latent condensing (LCON)}
\begin{figure}[hbtp]
\centering
\includegraphics[scale=0.35]{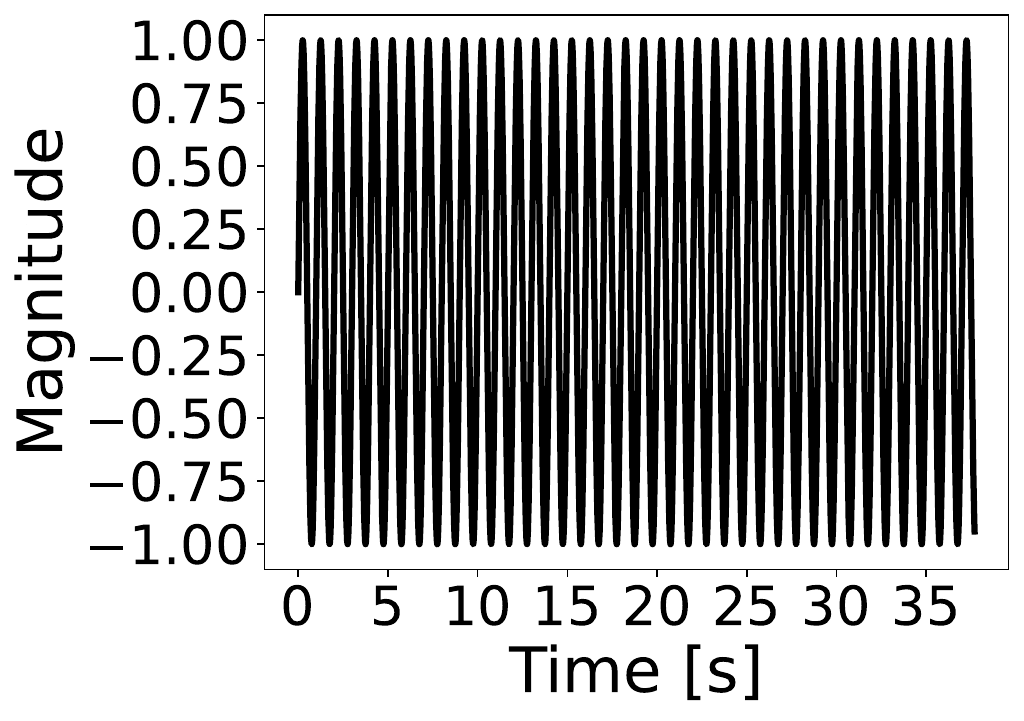} 
\includegraphics[scale=0.35]{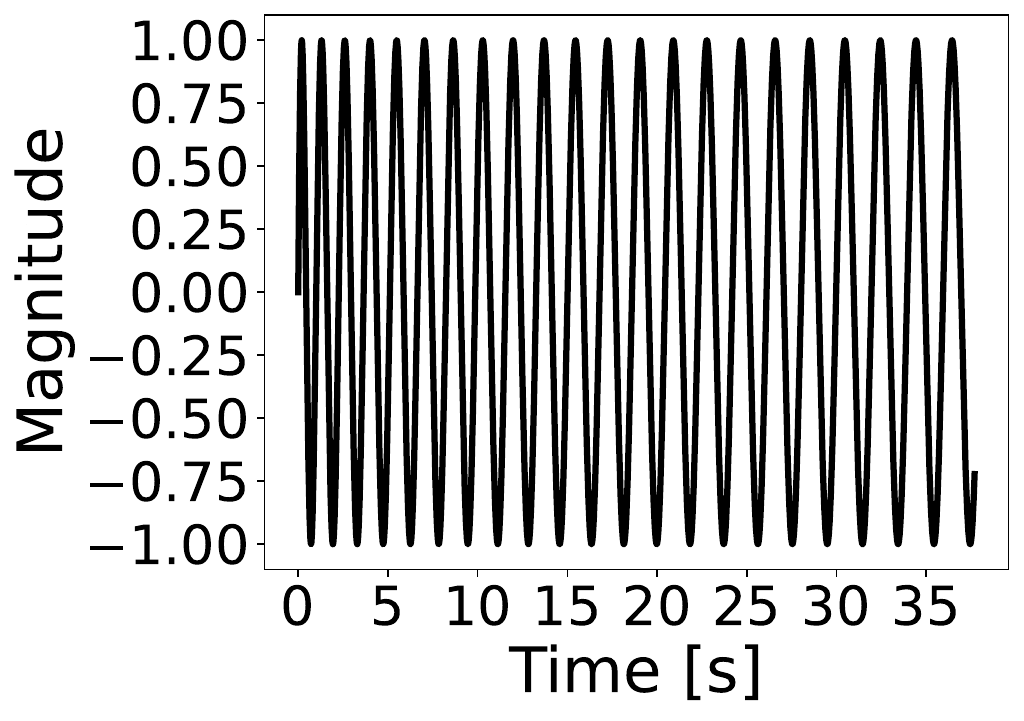} 
\caption{Two example signals, (left) $f(t) = sin(2\pi t)$, and (right) $f(t) = sin(2\pi t^{0.85})$ to illustrate
some of the functionality of LS-PIE.}\label{fig:ToyExamples}
\end{figure}

\begin{figure}[hbtp]
\centering
\includegraphics[scale=0.35]{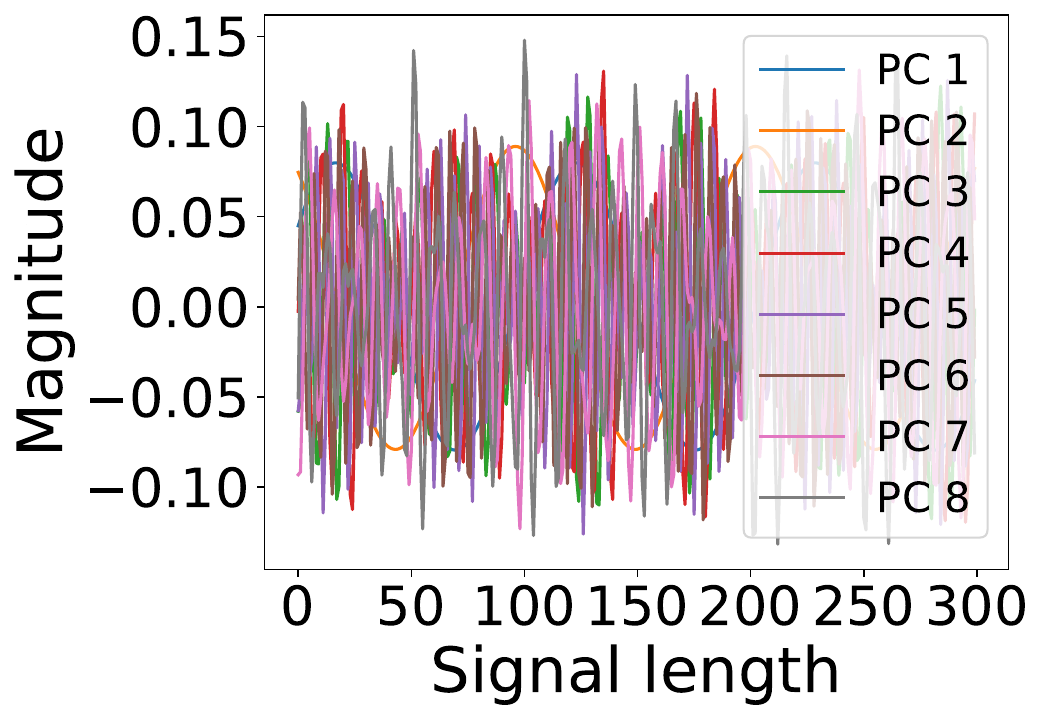} 
\includegraphics[scale=0.35]{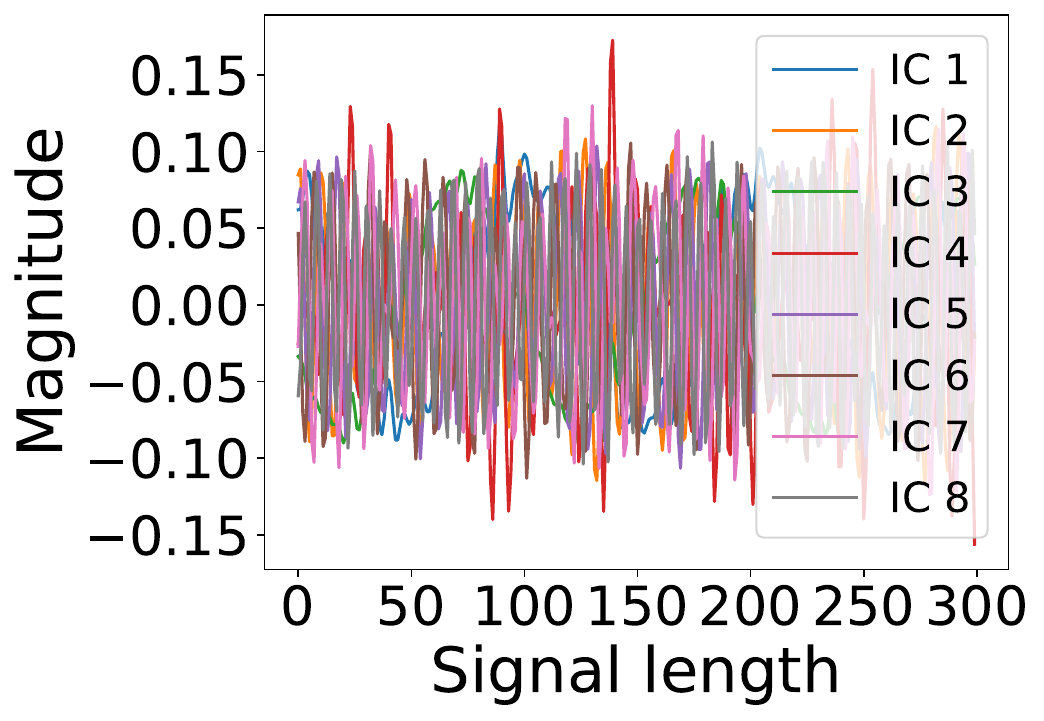} 
\includegraphics[scale=0.35]{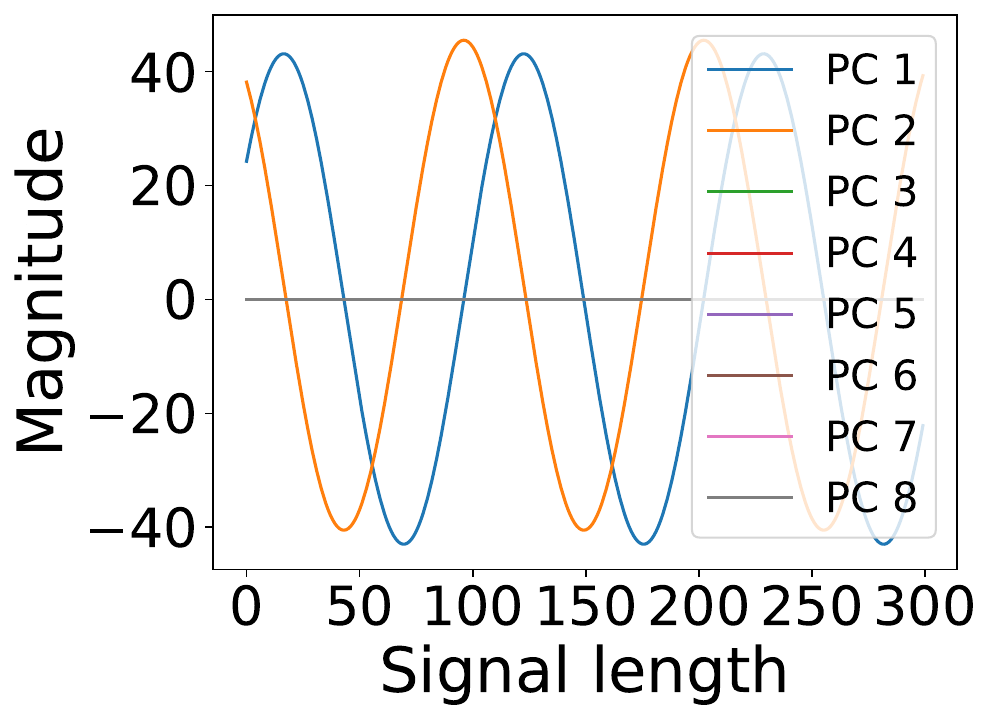} 
\includegraphics[scale=0.35]{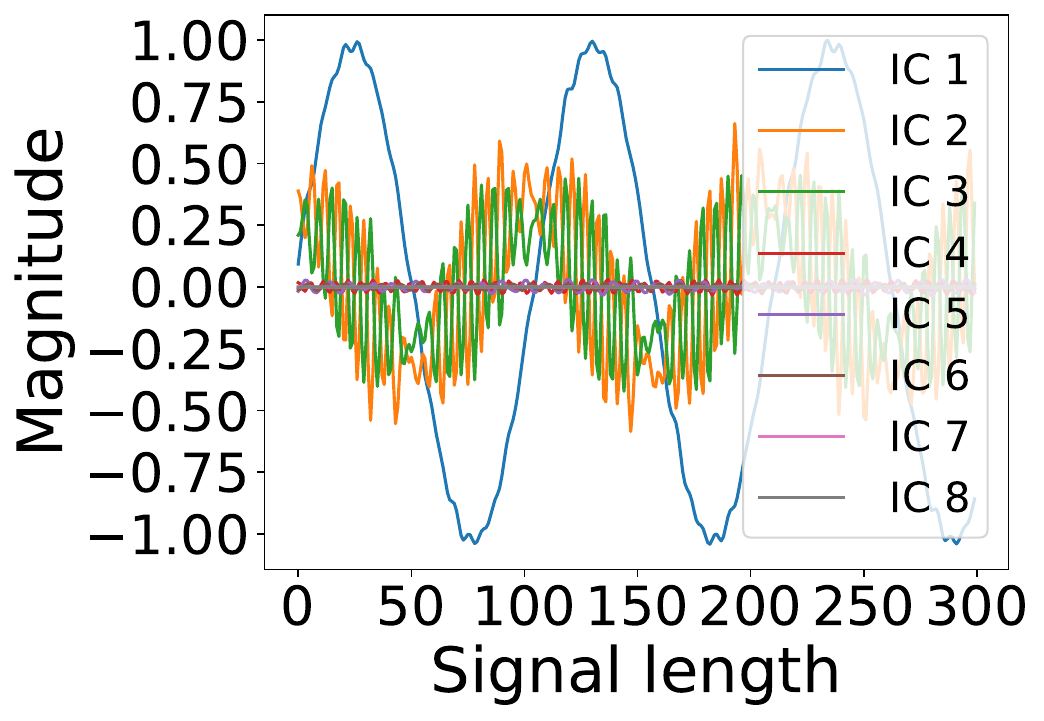} 
\includegraphics[scale=0.35]{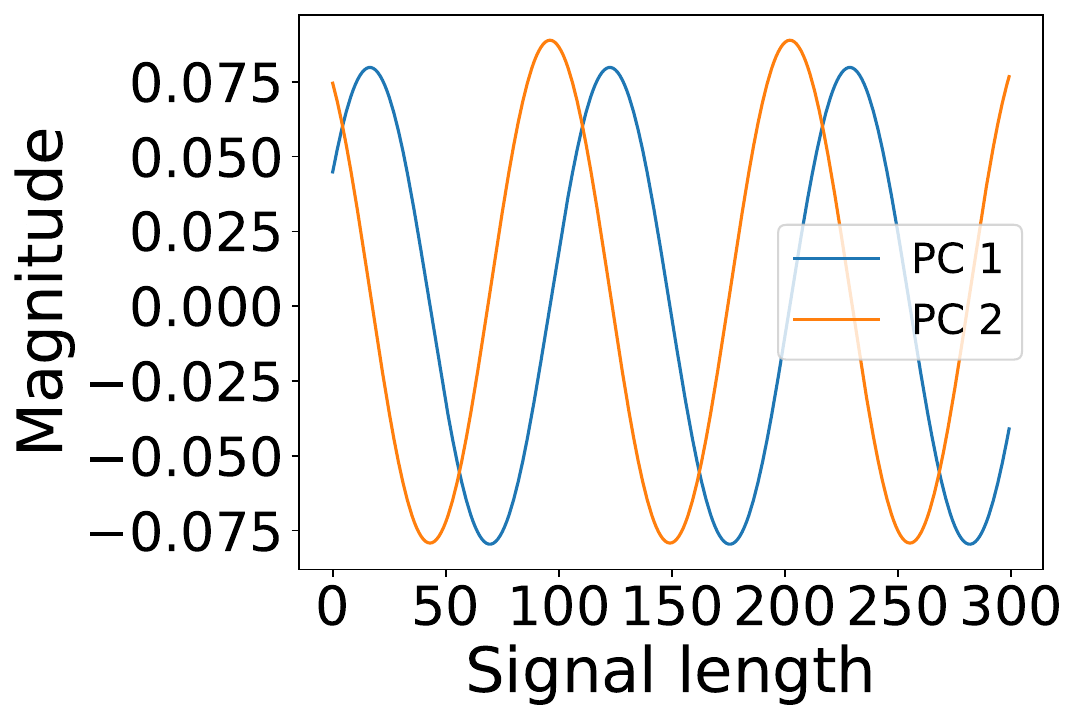} 
\includegraphics[scale=0.35]{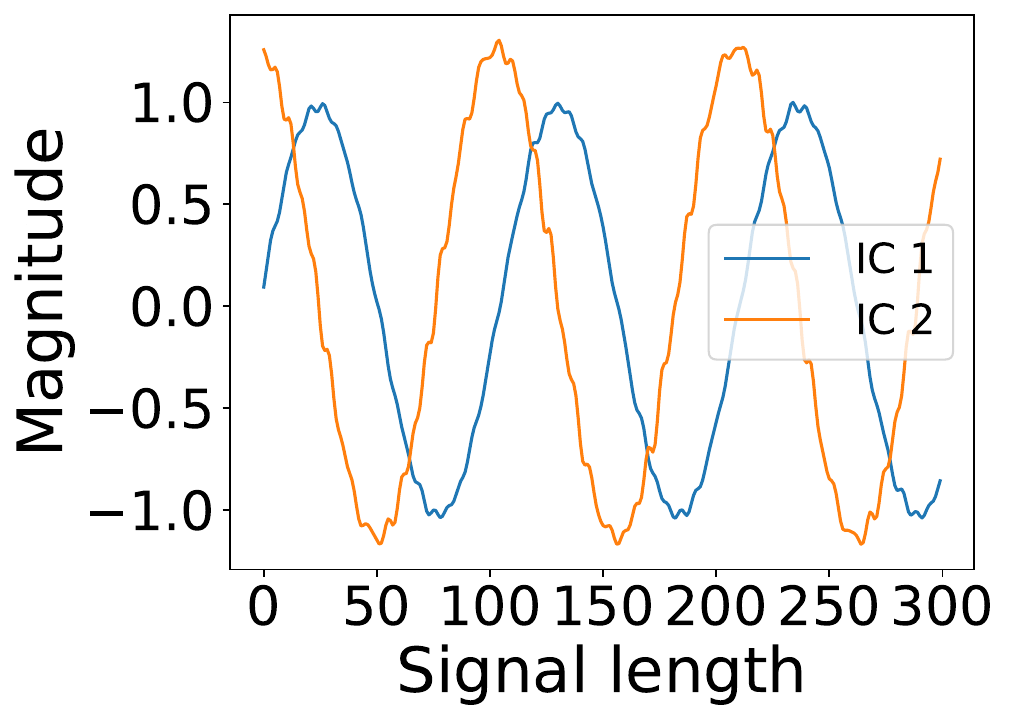} 

\caption{For the time series signal $f(t) = sin(2\pi t)$ depicting
 (top row) normalised latent directions for PCA and ICA, without applying
latent ranking (LR), or latent scaling (LS), (middle row)
variance-explained ranked and variance-explained scaled latent directions for PCA and ICA,
 (bottom row)
variance-explained ranked and variance-explained scaled latent directions with latent condensing (LCON) for PCA and ICA.
}\label{fig:Toy1}
\end{figure}

Consider the simplest example \[ f(t) = sin(2\pi t), \] uniformly sampled at $\frac{4000}{12\pi}$ samples per second using Hankelisation with a window length of 300. The results
for extracting 8 latent variables using PCA and ICA are shown in Figure~\ref{fig:ToyExamples}(left). Here, we expect identical results for PCA and ICA, merely a single-frequency Fourier sin-cosine decomposition as shown in Figures~\ref{fig:Toy1}. 
Note the improvement in informativeness as latent scaling (LS) are applied. In turn, note the improvement in the informativeness of the latent directions of latent ranking (LR) and enhancement of latent condensing (LCON) on ICA. For ICA, LC combined the
second and third ranked ICs.

\begin{figure}[hbtp]
\centering
\includegraphics[scale=0.35]{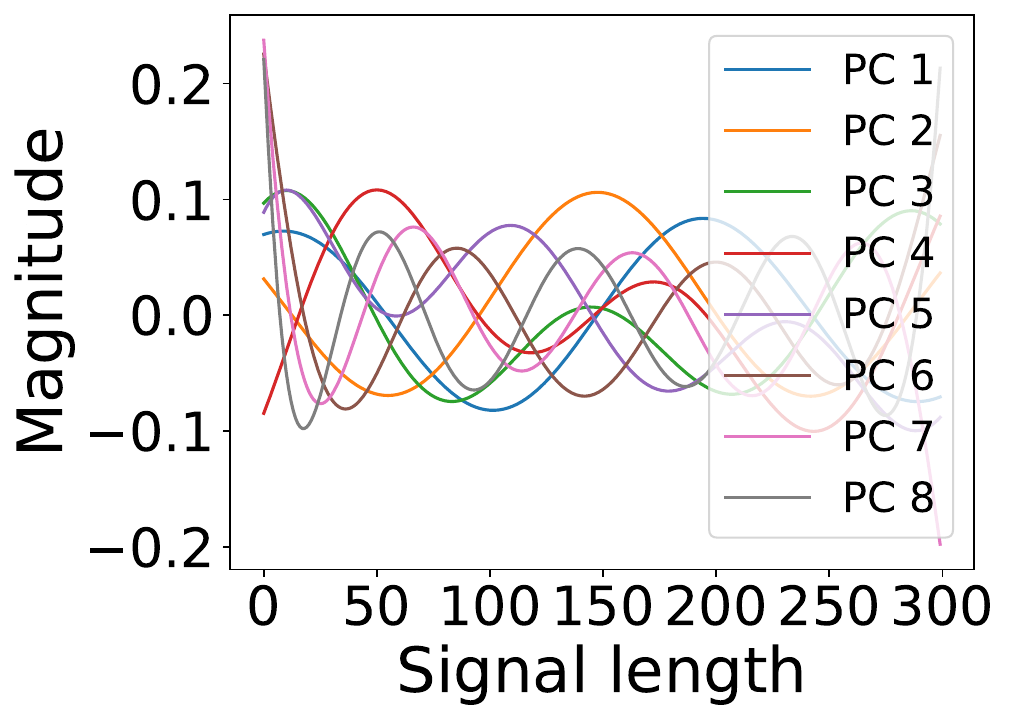} 
\includegraphics[scale=0.35]{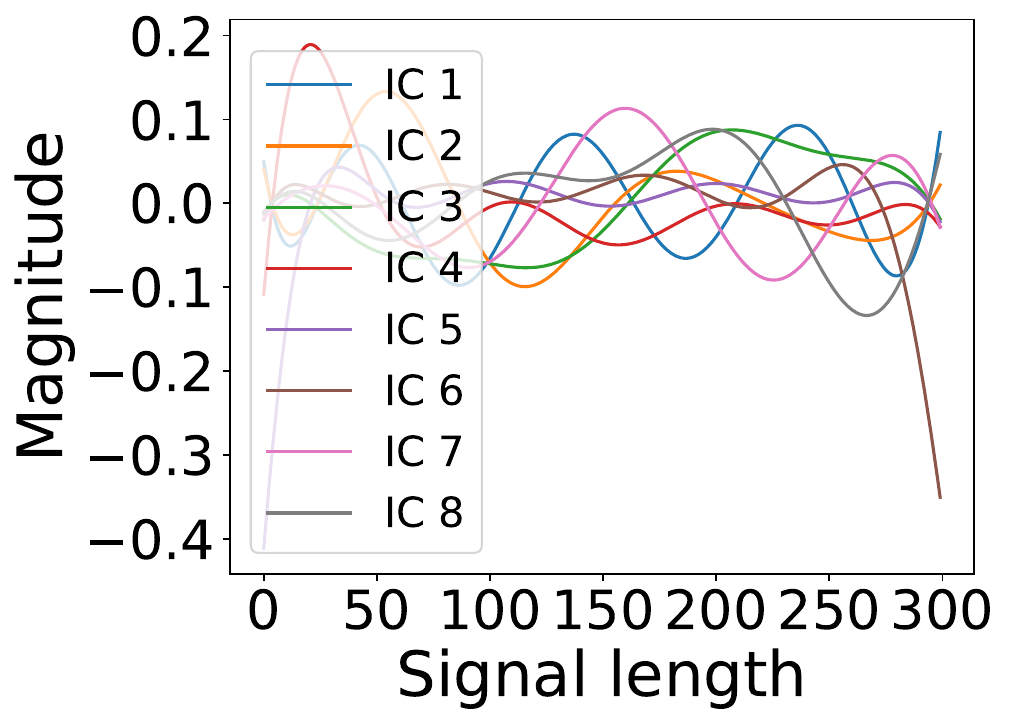} 
\includegraphics[scale=0.35]{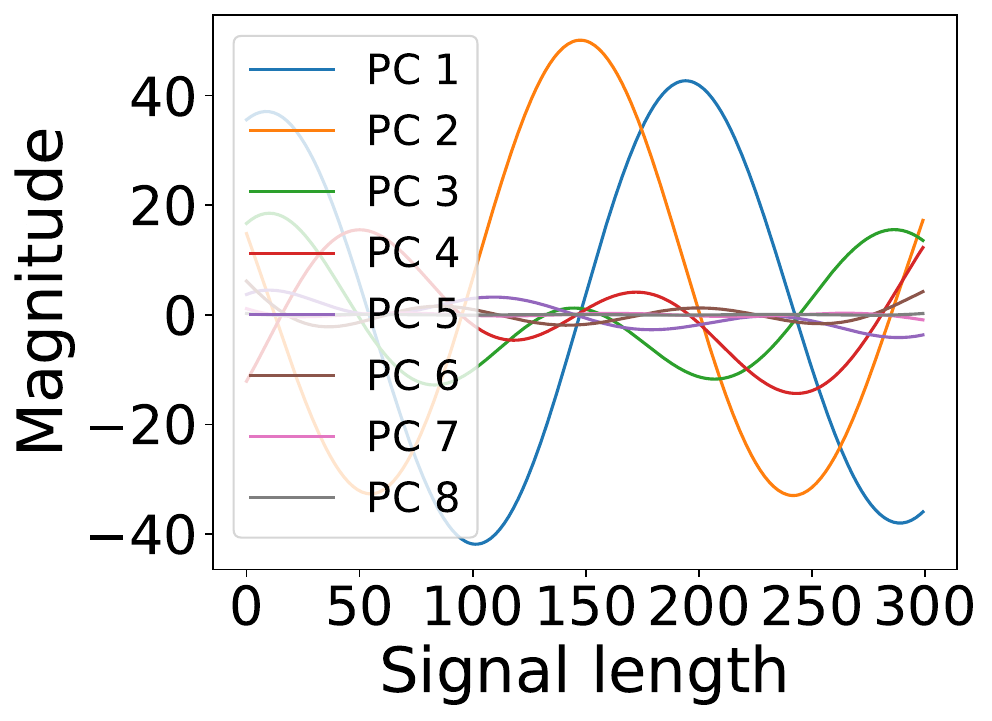} 
\includegraphics[scale=0.35]{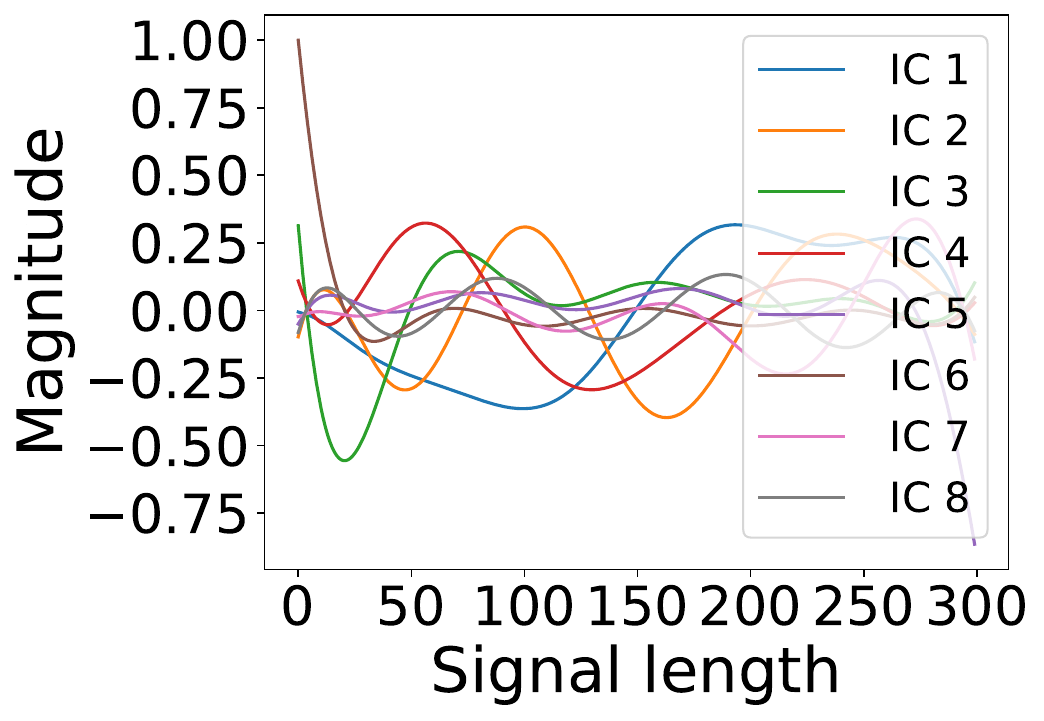} 
\includegraphics[scale=0.35]{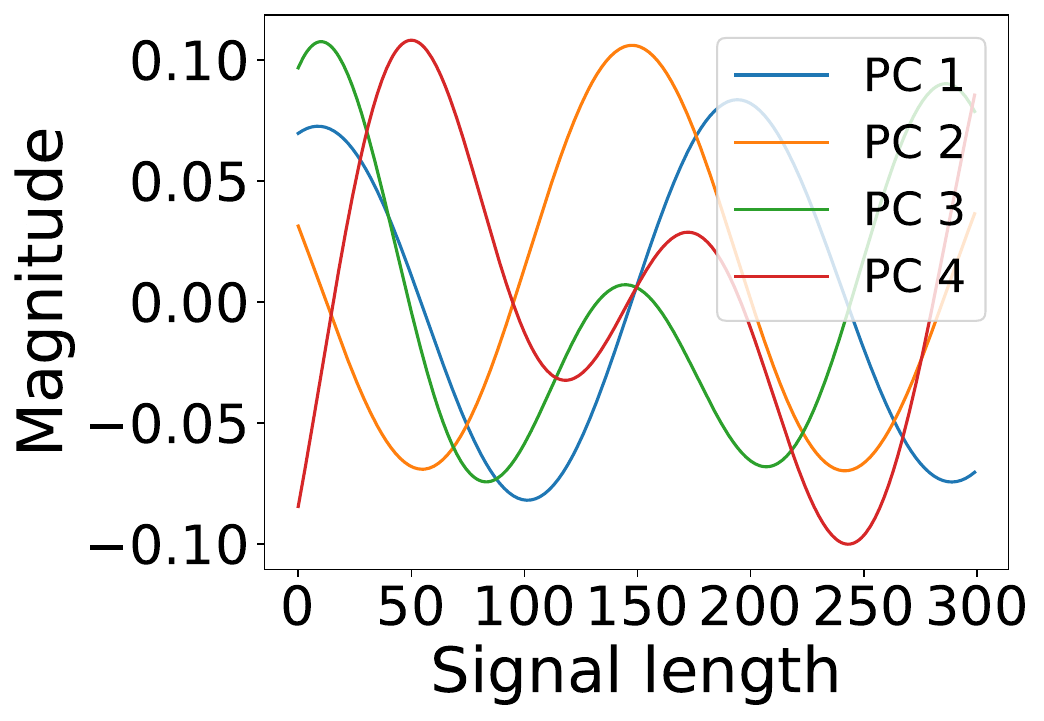} 
\includegraphics[scale=0.35]{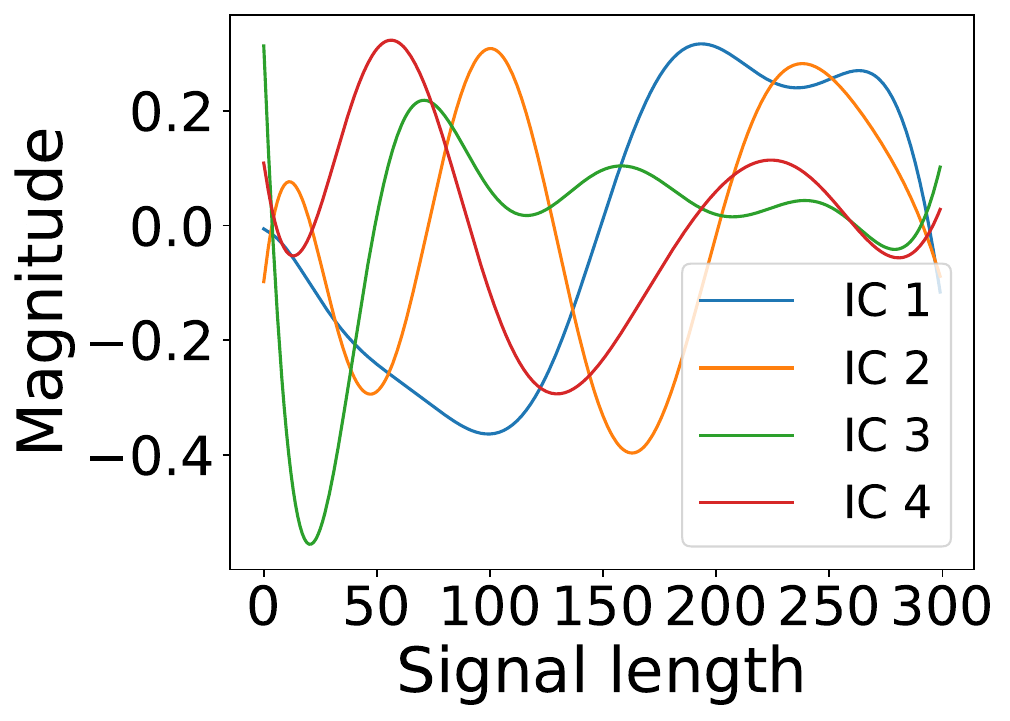}

\caption{For the time series signal $f(t) = sin(2\pi t^{0.85})$ depicting
 (top row) normalised latent directions for PCA and ICA, without applying
latent ranking (LR), or latent scaling (LS), (middle row)
variance-explained ranked and variance-explained scaled latent directions for PCA and ICA,
 (bottom row)
variance-explained ranked and variance-explained scaled latent directions with latent condensing (LCON) for PCA and ICA.}\label{fig:Toy2}
\end{figure}

In turn, Figure~\ref{fig:ToyExamples}(right)
is a signal with decreasing frequency over time, expressed by $f(t) = sin(2\pi t^{0.85})$. Here, we expect to see some differentiation in the latent directions between PCA and ICA as shown in Figures~\ref{fig:Toy2}. 
The improvement in interpretation and informativeness of the latent directions using LS-PIE is evident. LS-PIE isolates and enhances the essential latent directions that allow time for a critical interpretation of
the latent directions, and the comparison between LVMs.

\section{Impact}
\label{}

The role of LS-PIE in interrogating LVMs and enhancing latent directions is clearly demonstrated in
two foundational example problems. LS-PIE ensures that the user can focus their time and energy
on interpreting the latent space for latent inference, as opposed to first having to define an informative
latent space. The potential impact of LS-PIE is an improved adoption of interpretation-centered LVMs in signal processing,
vibration-based condition monitoring, actuarial sciences, finances, and social and physical sciences, as well as, the improved
interpretation of reconstruction-centered LVMs. 

This initial LS-PIE framework is a latent variable ecosystem to enhance the 
practical application of LVMs and centering research activity of LVMs around latent inference for
interpretation.

\section{Conclusions}
\label{}

LS-PIE improves the interpretability of reconstruction-centered and in-terpretation-centered LVMs through latent ranking and latent scaling, while enhancing the information spread over latent directions
through latent condensing. Two foundational datasets clearly highlight the benefit of utilising LS-PIE to enhance the informativeness of reconstruction-centered and interpretation-centered LVMs.

The LS-PIE framework is the first step towards an LVM ecosystem that benefits the practical application and research opportunities of LVMs. Future research will develop additional functionality
that benefits LVM research and practical deployment of LVM for industrial applications.

\section{Conflict of Interest}
%Please select the appropriate text:

No conflict of interest exists:
We wish to confirm that there are no known conflicts of interest associated with this publication and there has been no significant financial support for this work that could have influenced its outcome.

%\section*{Acknowledgements}
%\label{}
%
%All parties that contrOptionally thank people and institutes you need to acknowledge. \\
%Eyyyy look here \cite{Bach2003} These citations will maybe work \cite{Zhao2009} \citep{DiMauro2022}



%% References:
\bibliographystyle{elsarticle-num} 
\bibliography{lib.bib}




%%\bibitem{}

%%\end{thebibliography}
%Please add the reference to the software repository if DOI for software  is available. 

%\section*{Current executable software version}
%\label{}
%
%Ancillary data table required for sub version of the executable software: (x.1, x.2 etc.) kindly replace examples in right column with the correct information about your executables, and leave the left column as it is.
%
%\begin{table}[!h]
%\begin{tabular}{|l|p{6.5cm}|p{6.5cm}|}
%\hline
%\textbf{Nr.} & \textbf{(Executable) software metadata description} & \textbf{Please fill in this column} \\
%\hline
%S1 & Current software version & 1.0 \\
%\hline
%S2 & Permanent link to executables of this version  & For example: $https://github.com/combogenomics/$ $DuctApe/releases/tag/DuctApe-0.16.4$ \\
%\hline
%S3 & Legal Software License & List one of the approved licenses (Look at list of approved Licenses) \\
%\hline
%S4 & Computing platforms/Operating Systems & Windows, macOS,  Linux \\
%\hline
%S5 & Installation requirements \& dependencies & \\
%\hline
%S6 & If available, link to user manual - if formally published include a reference to the publication in the reference list & For example: $http://mozart.github.io/documentation/$ \\
%\hline
%S7 & Support email for questions & u16301545@up.ac.za\\
%\hline
%\end{tabular}
%\caption{Software metadata (optional)}
%\label{} 
%\end{table}

\end{document}